\documentclass[letterpaper, 10 pt, conference]{ieeeconf}  %

\IEEEoverridecommandlockouts                              %

\usepackage{graphics} %
\usepackage{epsfig} %
\usepackage{multirow}
\usepackage{float}
\usepackage{amsmath}
\usepackage{amssymb}
\usepackage{diagbox}
\usepackage{subfig}
\usepackage{xcolor}
\newcommand{\taninv}{\tan^{-1}}
\usepackage{cancel}

\title{\LARGE \bf
Flash: Fast and Light Motion Prediction for Autonomous Driving with Bayesian Inverse Planning and Learned Motion Profiles}

\author{Morris Antonello$^{1*}$, Mihai Dobre$^{1*}$, Stefano V. Albrecht$^{1, 2}$, John Redford$^{1}$ and Subramanian Ramamoorthy$^{1, 2}$%
\thanks{$^*$ Equal contribution}%
\thanks{$^{1}$Applied Research Team, Five AI, Edinburgh, United Kingdom
        {\tt\small \{morris.antonello, mihai.dobre\}@five.ai}}%
\thanks{$^{2}$School of Informatics, University of Edinburgh, Edinburgh, United Kingdom}
}%

\begin{document}

\maketitle
\thispagestyle{empty}
\pagestyle{empty}

\begin{abstract}

Motion prediction of road users in traffic scenes is critical for autonomous driving systems that must take safe and robust decisions in complex dynamic environments. We present a novel motion prediction system for autonomous driving. Our system is based on the Bayesian inverse planning framework, which efficiently orchestrates map-based goal extraction, a classical control-based trajectory generator and a mixture of experts collection of light-weight neural networks specialised in motion profile prediction. In contrast to many alternative methods, this modularity helps isolate performance factors and better interpret results, without compromising performance. This system addresses multiple aspects of interest, namely multi-modality, motion profile uncertainty and trajectory physical feasibility. We report on several experiments with the popular highway dataset NGSIM, demonstrating state-of-the-art performance in terms of trajectory error. We also perform a detailed analysis of our system's components, along with experiments that stratify the data based on behaviours, such as change-lane versus follow-lane, to provide insights into the challenges in this domain. Finally, we present a qualitative analysis to show other benefits of our approach, such as the ability to interpret the outputs.
\end{abstract}

\section{INTRODUCTION}

Motion prediction is receiving significant attention because of the need for reactive decision making in numerous robotics applications. In autonomous driving, it is the problem of inferring the future states of road users in the traffic scene. Navigating complex dynamic environments in a safe and efficient manner requires observing, reasoning about and responding to relevant objects or hazards along the way. Complex reasoning~\cite{hubmann2018, albrecht2021interpretable, eiras2021two, rhinehart2021}, such as when to perform maneuvers or low-level path planning optimisations, is not possible without making predictions about various properties of interest, such as agent future goals and trajectories.

Motion prediction is challenging because of the multitude of factors that determine future behaviour, not all of which may be explicitly modelled. Algorithms are expected to generalise to different environmental situations, such as a wide range of different layouts including merges, junctions, roundabouts etc. Combined with varying intentions and behaviours of different agents, this can result in many possible future outcomes that need to be captured. Furthermore, low-level aspects that affect the possible future trajectories, such as vehicle characteristics (e.g. dimensions, turning circle etc), type of vehicle (e.g. emergency services, transport, motorbikes etc) and driving styles (e.g. aggressive versus conservative), further increase the resulting uncertainty.
Besides multi-modality and spatial uncertainty, using motion prediction within a safety critical and real-time motion planning system extends the list of requirements. System maintainability and physical realism are additional qualities of interest which are less frequently addressed~\cite{girase2021physically}, e.g. many state-of-the-art methods, such as~\cite{deo2018convolutional, mercat2020multi, song2020pip, mersch2021maneuver}, are end-to-end regression-based systems which might not produce physically-feasible trajectories. As described in~\cite{sculley2014machine}, machine learning systems can incur massive maintenance costs because of system-level anti-patterns, such as entanglement which prevents isolated improvements. 

In contrast, hybrid systems, such as our proposed motion prediction system, allow including light-weight machine learning models, isolate performance factors, replace components and interpret results while demonstrating state-of-the-art performance. Our proposed prediction system is general and can address the majority of the previously described challenges. In this paper, we evaluate our method in the highways setting, which defines a microcosm of the complex dynamics that one expects to find in everyday driving~\cite{srinivasan2021comparing}. Highways include multiple lanes;
space-sharing conflicts are common between the on-ramp vehicles and vehicles on the outermost lane and, similarly, during change-lane and overtaking behaviours when vehicles need finding suitable gaps while maintaining safety distances.

Our contributions are two-fold:
\begin{itemize}
    \item we present Flash, a novel hybrid motion prediction system that is based on the Bayesian inverse planning framework. It efficiently orchestrates map-based goal extraction, a classical control-based trajectory generator and a mixture of experts collection of light-weight neural networks specialised in motion profile prediction. This system models properties of interest such as multi-modality and motion profile uncertainty while providing strong guarantees that kinematic and motion profile constraints are respected;
    \item we evaluate the system thoroughly on the popular highway dataset - NGSIM~\cite{colyar2007us101}, comparing with alternative multimodal methods. Flash improves the state-of-the-art trajectory error reported in~\cite{mercat2020multi} by 9.05\,\% for a 5\,s prediction horizon, it guarantees kinematic and motion profile constraints by construction, and its training is 96.92\,\% faster than in~\cite{mersch2021maneuver}. We analyse single components such as motion profile prediction and Bayesian inference, showing that the modularity helps isolate performance factors and interpret predictions.
\end{itemize}

\section{RELATED WORK}
\label{related_works}
The proposed system predicts the motion of the traffic participants with a Bayesian inverse planning approach~\cite{baker2009action, ramirez2010}. Previous work~\cite{albrecht2021interpretable, hanna2021} has shown that Bayesian inference combined with defined behaviours extracted from a layout of the scene generates predictions that are explainable by means of rationality; i.e. optimality given certain metrics. Another benefit of such an architecture, where Bayesian inference is the top level component, is its efficiency as reported by \cite{luo2022gamma}; we also show empirically this is true.

Planning literature has a long history of using vehicle models in combination with low-level trajectory generators, see \cite{gonzalez2016} for a review of algorithms for generating paths, motion profiles and trajectories. On the other hand, such methods are rarely used in prediction. \cite{buhet2020} have combined a data-driven method with a polynomial to offer some level of smoothness in the output. This approach doesn't offer any guarantees, and the resulting trajectories can still be kinematically infeasible. To address this limitation, \cite{girase2021physically} have used a vehicle model together with a path following algorithm, in particular pure-pursuit~\cite{coulter1992implementation}. We use a similar solution in our trajectory generation component.

There are several examples of prediction methods that focus on a simplified version of the problem: motion profile prediction, see \cite{lefevre2014comparison} for a comparison. In this space, physics-based models, e.g. constant velocity and constant acceleration, can be accurate in the short-term, but they do not consider other traffic agents. More evolved analytical solutions, such as Intelligent Driver Model (IDM)~\cite{treiber2000congested} to generate a car following behaviour and MOBIL~\cite{kesting2007general} for change-lane behaviour, have been used successfully in planning~\cite{hubmann2018} and simulation for testing~\cite{bernhard2021}. Still, these are deterministic methods and limited to capture only part of the context. To address these limitations, we integrate Bayesian inverse planning with a collection of expert neural networks, in particular mixture density networks~\cite{bishop1994mixture} modelling motion profile uncertainty~\cite{ mercat2020multi, mercat2019kinematic}, to predict the future motion profile while considering traffic context, i.e. the motion and relative spatial configuration of neighbouring agents.

Neural network based approaches have become very popular in motion prediction. Initially, these data-driven methods have been proposed without the use of maps~\cite{alahi2016}. However, utilising a map can have benefits, for example anchoring to the driveable area, capturing rules of the road, reducing the hypothesis space, disambiguate intentions etc. Lately, many high-performing works in structured environments rely on a map in a variety of ways: from minor cues~\cite{deo2018convolutional, song2020pip, mersch2021maneuver} to a rasterised version that contains the complete information including traffic signals in some cases~\cite{chai2019, gao2020, chou2020, zhang2020}. Another categorisation criterion is that these implementations are generally end-to-end trained. Our system is also heavily reliant on a map definition. Different to previous work, our networks do not consume map information at input, but rather are trained as specialised on behaviours extracted from the layout which can be reused when that behaviour is available. Such a specialisation permits independent training, tuning and inspection.

\section{PROBLEM DEFINITION}
The objective of motion prediction is to produce possible future trajectories and estimate how likely these are given the history of observations of the observable agents. We define the history $\mathbf{h}^i$ of $k + 1$ observations for an agent $i$ as a sequence of coordinates $(x,y)$, orientations $\theta$, and velocity vectors $\mathbf{v} \in \mathbb{R}^2$: $\mathbf{h}^i = [(x^i_t, y^i_t, \theta^i_t, \mathbf{v}^i_t)]^{k}_{t=0}$, preceding and containing the current time step $t=k$. Similarly, we define the predicted $\hat{\mathbf{y}}^i = [(\hat{x}^i_t, \hat{y}^i_t, \hat{\theta}^i_t, \hat{\mathbf{v}}^i_t )]^T_{t=k+1}$ and ground truth future $\mathbf{y}^i = [(x^i_t, y^i_t, \theta^i_t, \mathbf{v}^i_t)]^T_{t=k+1}$ trajectories up to the horizon $T$. These can be decomposed into a path, which is a sequence of $N$ positions $[(x^i, y^i)]^N_{n=0}$, and a motion profile, which is a sequence of speeds $\mathbf{s}^i = [s^i_t]^T_{t=k+1}$ from which higher-order derivatives such as acceleration $a^i_t$ and jerk $j^i_t$ can be estimated.
There are two challenges that a prediction system faces: the space of possible future trajectories is continuous, and the future is uncertain. We handle the former challenge by predicting a spatial uncertainty attached to the discrete predicted trajectories. We model this spatial uncertainty with a multivariate Gaussian capturing the position variance at each predicted future state.
The latter challenge is addressed in our system by producing a multimodal output, i.e. a discrete distribution over a set of predicted trajectories $P(\hat{\mathbf{y}}^i)$.

\section{METHODS}
\label{methods}

\subsection{Overview} 
\begin{figure*}[ht]
  \centering
  \includegraphics[width=0.65\linewidth]{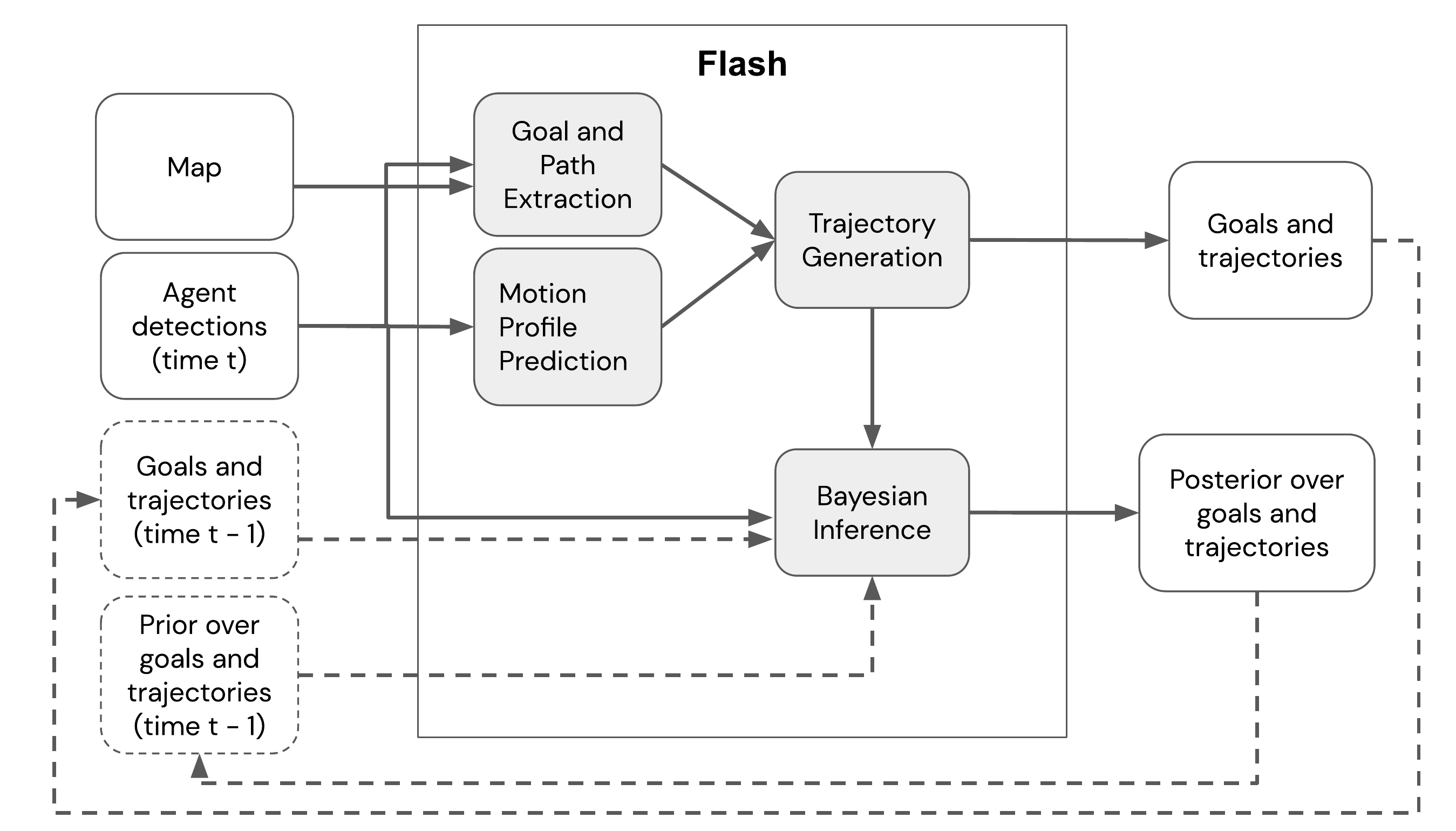}
  \caption{System overview. The diagram shows the main components in grey boxes and their interconnections. Taking a Bayesian inverse planning approach, outputs become inputs at the next time step as shown by the dotted lines.}
  \label{system_overview}
\end{figure*}

The proposed model performs multimodal prediction by taking a Bayesian inverse planning approach~\cite{baker2009action}; it recursively compares previously predicted trajectories with observations to estimate likelihoods and computes a joint posterior distribution over goals and trajectories using Bayesian inference. A system overview is shown in Figure~\ref{system_overview}. It involves four main components which we will discuss in more detail: \textit{i.)} goal and path extraction, \textit{ii.)} motion profile prediction, \textit{iii.)} trajectory generation and \textit{iv.)} Bayesian inference.

\subsection{Goal and Path Extraction}
\begin{figure}[t]
  \centering
  \includegraphics[scale=0.45]{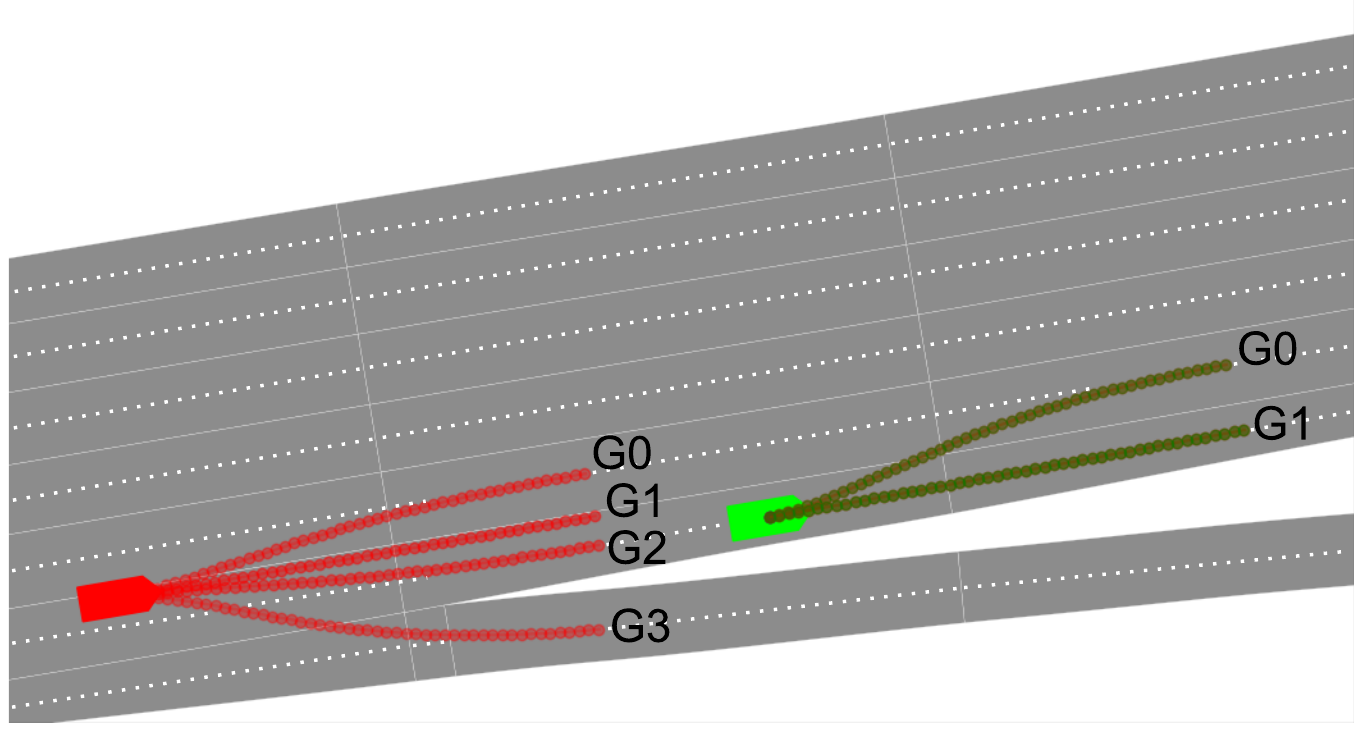}
  \caption{Portion of the I-80 highway layout showing the lane centre lines (dotted white lines) used as target paths for trajectory generation, annotated hypothesised goals and trajectories (coloured lines) for a single agent at different time steps (different colour). The agent's hypothesised goals depend on its position on the road layout.}
  \label{fig:goal_path_extraction}
\end{figure}

High-definition maps are useful in various aspects of self-driving; for example a map can help disambiguate the intentions of other agents~\cite{albrecht2021interpretable, hanna2021,brewitt2021grit} and aid planning in challenging traffic situations~\cite{shwartz2017}. We follow the OpenDrive standard~\cite{dupuis2010opendrive} and implement our own layout definition for querying the geometry of roads and lanes, their properties, e.g. lane types, and how they are interconnected. Given the position and orientation of an agent $i$ at time $t$, we extract its possible goals $\mathbf{g}^i_t$ by exploring all lane graph traversals up to a depth. In highway situations, the immediate goals correspond to staying in the lane, or staying in the lane while maintaining the current lateral offset to the centre of the lane, or changing to a neighbour lane, or entering the highway if the agent is on the entry ramp, or exiting it if the agent is on the slow lane close to an exit ramp. We refer to this set of goals as the \textit{hypothesised goals}. Figure~\ref{fig:goal_path_extraction} illustrates the use of lane centre lines with an optional offset as target paths for trajectory generation. The goals correspond to intentions that an agent could have and our system multimodality derives from this set of possible intentions.

\subsection{Motion Profile Prediction}
The predictive performance of different motion profile models varies strongly with the prediction horizon and traffic conditions, such as the congestion level~\cite{lefevre2014comparison}. We chose a neural network-based approach for this task as these capture more contextual variations and exploit the availability of training data~\cite{lefevre2014comparison}. Due to our proposed architecture, the motion profile prediction task is narrow and well-defined. In addition to this, we also further split the data according to the behaviour being performed, i.e. follow-lane versus change-lane, and context, i.e. number of front and side agents. These aspects permit the use of a mixture of experts~\cite{rokach2010pattern} collection of specialised and lightweight Mixture Density Networks (MDNs)~\cite{bishop1994mixture}, each focusing on a different subtask.

Each expert model in the collection consumes a different sized 1D feature vector $\mathbf{z}$. The features capture properties of the target agent being predicted as well as properties of front agents and side agents in the target agent's neighbourhood given the current state only. We assume the data has been filtered using an object tracker. The properties include agents' speeds, acceleration values, the agent class $c \in \{car, truck, motorbike, ...\}$. Follow-lane models include the headway distances from the target vehicle. Change-lane models consume additional features such as the neighbouring side, if the side agents are in front or behind the target vehicle, the centre to centre distance to the target vehicle and the shortest distance between the vehicles' polygons and the target vehicle's polygon. The input for the change-lane models contains only features of the agents on the neighbouring side that the change-lane is performed towards, i.e. left vs right. Highway entries and exits are considered change-lanes.
In this work, we conduct experiments with agents within a 60\,m radius as in~\cite{mercat2020multi}, including up to 3 agents in front of target vehicle and up to 3 side agents on each side. Given this, the final collection of experts is made of 4 follow-lane models, each specialised for a different number of front agents, and 16 change-lane models, each specialised for a different number of front and side agents. At runtime, multiple models are selected based on the extracted goals, which can correspond to follow-lanes or change-lanes, and the number of surrounding agents.

The neural networks' objective is to predict distances from current position over the next $\tilde{N}$ time steps, so the target is a sequence of distances $\mathbf{y}^d_{t=0} = [r_t]_{t=1}^{\tilde{N}}$. We define the predicted sequence of distances as $\hat{\mathbf{y}}^{d}_{t=0}$. $\tilde{N}$ is a function of horizon and delta time, respectively 5\,s and 1\,s, hence 5. Each MDN model learns the joint distribution $p(\mathbf{z}, \mathbf{y}^d)$ in the form of a multivariate Gaussian mixture model with $M$ multivariate Gaussian functions:
\begin{equation}
\label{eq:model_output}
\begin{aligned}
p(\mathbf{z}, \mathbf{y}^d) = \sum_{m = 1}^{M} \pi_m \mathcal{N}(\mathbf{z}, \mathbf{y}^d | \boldsymbol{\mu}_m, \Sigma_m) 
\end{aligned}
\text{, where}
\end{equation}
$\pi_m$, $\boldsymbol{\mu}_m$ and $\Sigma_m$ are the probability, mean vector, and diagonal covariance matrix of the $m^{th}$ multivariate Gaussian mixture component. MDNs also model uncertainty, predicting the mixture parameters including variances instead of single outputs. We denote each $m$ component's prediction error as $\boldsymbol{\epsilon}_m$ and train the model with the Negative Log Likelihood (NLL) loss function:
\begin{equation}
\label{nll_loss}
\begin{aligned}
NLL = -\ln{\sum_{m = 1}^{M} \pi_m e^{-\frac{1}{2} \boldsymbol{\epsilon}^T_m \Sigma^{-1}_m \boldsymbol{\epsilon}_m - \ln{(\sqrt{(2\pi)^{\tilde{N}}|\Sigma_m|})}}}
\end{aligned}
\end{equation}

In this work, we set $M$ to 1 since using larger values did not impact the overall system performance. Indeed, we rely on goal extraction and the collection of experts to handle multimodality.
The component's vector of predicted means $\boldsymbol{\mu}_m$ is used to construct the predicted motion profile $\hat{\mathbf{s}}^i$ for an agent $i$ given the known delta time of 1\,s.
In this architecture, each MDN consists of 2 fully connected layers, 64 and 32 neurons each, with $relu$ activations. Each model is trained independently on a specialised dataset split, augmented with samples from the other splits. For instance, follow-lane networks considering two front agents are trained with follow-lane samples with at least 2 front agents and, similarly, change-lane networks considering two side agents are trained with change-lane samples with at least 2 side agents. We tuned the hyper-parameters using randomised search on the training set. Follow-lane networks are trained with a learning rate of 0.001 and a batch size of 1024 for 10 epochs. Change-lane networks are trained with the same learning rate and a batch size of 32 for 20 epochs.

\subsection{Pure Pursuit Trajectory Generator}
\label{pure-pursuit}
\begin{figure}[t]
  \centering
  \includegraphics[width=0.65\columnwidth]{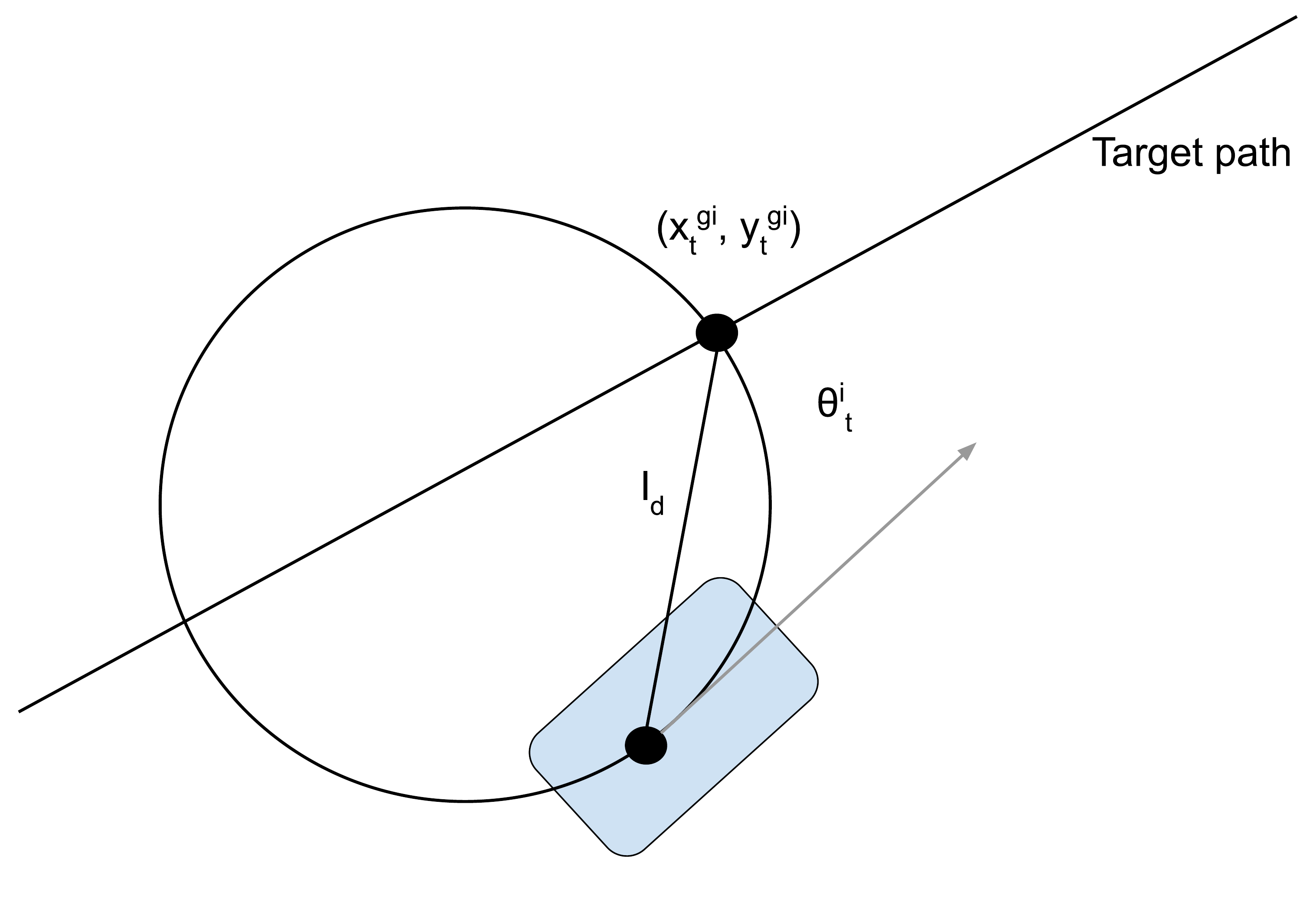}
  \caption{Goal selection for pure pursuit trajectory generator.}
  \label{fig:pure_pursuit}
\end{figure}

\begin{figure}[t]
  \centering
  \includegraphics[width=0.65\columnwidth]{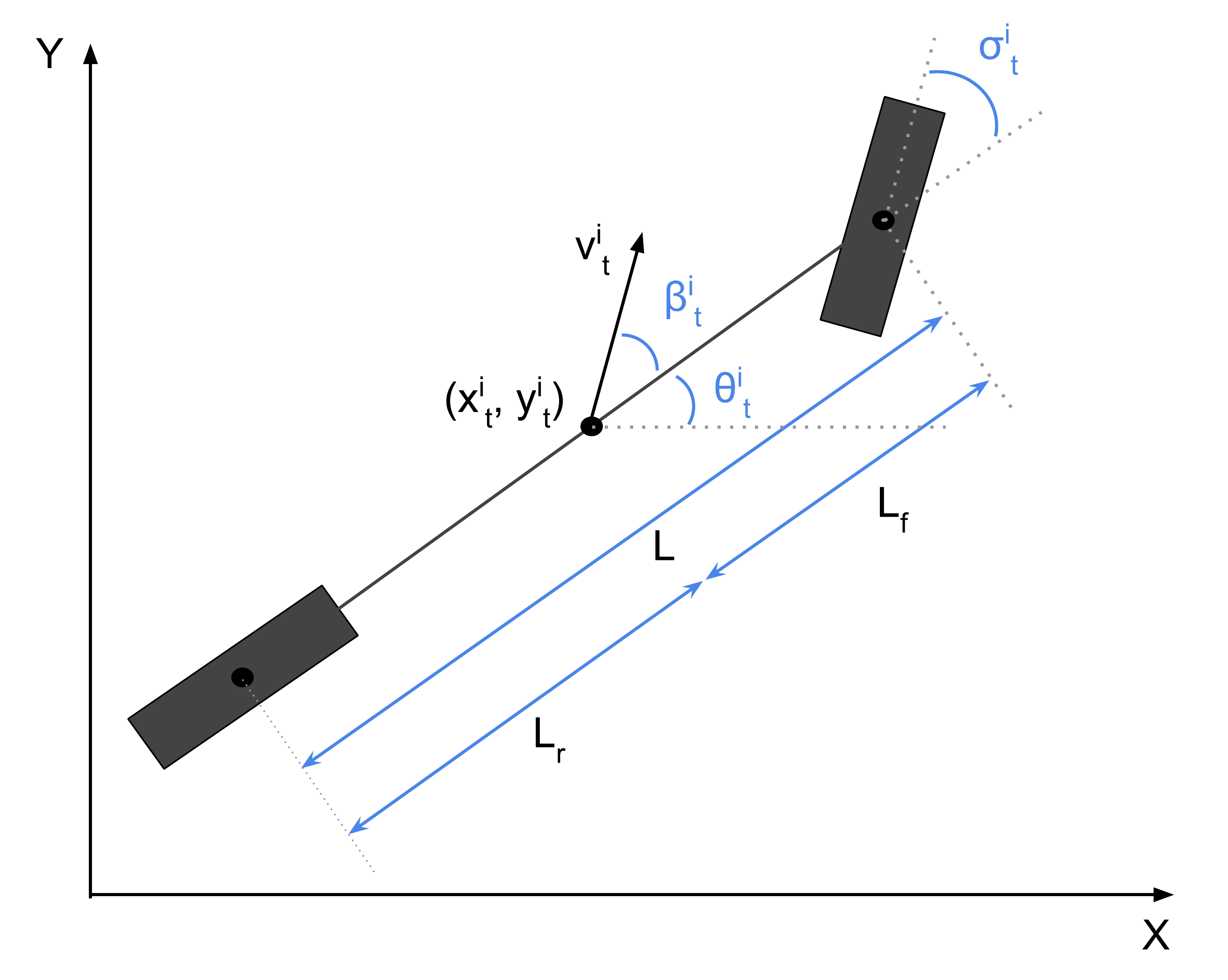}
  \caption{Kinematic bicycle model.}
  \label{fig:bicycle_model}
\end{figure}

Despite their strong ability to model context, neural networks are not interpretable and can perform poorly when conditions change~\cite{angelos2020, pulver2020}. Similarly to previous work~\cite{girase2021physically}, we use pure pursuit~\cite{coulter1992implementation} to address these challenges and generate physically feasible trajectories describing how each agent might reach any of its hypothesised goals. Pure pursuit is a path tracking algorithm that uses a controller and a vehicle model to generate a trajectory that minimises the error from the target motion profile and the target path. In our implementation, the neural networks provide the target motion profile while the target path is extracted from the map as previously described. Intuitively, the path tracking approach imitates how humans drive; they look at a point they want to go to and control the car to reach it. As shown in Figure~\ref{fig:pure_pursuit}, the algorithm chooses a goal position $(x^{g_i}_t, y^{g_i}_t)$ using a lookahead distance parameter $l_d$ and calculates the curvature that will move an agent from its current rear axle position to the target position while maintaining a fixed steering angle. We fixed $l_d$ to 10.0\,m in our experiments. This parameter is configurable and could account for current speed or target path curvature. 
An agent's state at time $t$ is described using its centre position $(x^i_t, y^i_t)$, current orientation $\theta^i_t$ and speed $s^i_t$. All agents are vehicles in the explored dataset, so we chose a kinematic bicycle model to represent their motion. In addition to the state of the vehicle, the bicycle model also requires the distance between the rear axle and the centre position $L_r^i$ and the wheelbase length $L^i$. Given a control input $u^i_t = (a^i_t, \sigma^i_t)$ at time $t$ composed of acceleration and steering angle respectively, the Equation~\ref{eq:bicycle_model} and Figure~\ref{fig:bicycle_model} describe the motion of the vehicle. The side slip angle $\beta^i_t = \taninv(\frac{L_r^i}{L^i} \tan(\sigma^i_t))$ needs to be computed given that the centre position of the agent is used. Finally, ${\Delta t}$ represents the time difference between two time steps.
\begin{equation}
\label{eq:bicycle_model}
\begin{aligned}
d&= s^i_t {\Delta t} + \frac{a^i_t {\Delta t}^2}{2} \\
x^i_{t+1}&= x^i_t + d \cos(\theta^i_t + \beta^i_t) \\
y^i_{t+1}&= y^i_t + d \sin(\theta^i_t + \beta^i_t) \\
\theta^i_{t+1}&= \theta^i_t + \frac{d}{L} \cos(\beta^i_t) \tan(\sigma^i_t) \\
s^i_{t+1}&= s^i_t + a^i_t {\Delta t}
\end{aligned}
\end{equation}

The control input $u^i_t$ is calculated using a proportional controller with the same gain $k_p=2$ for both the steer and acceleration components. The acceleration input at time $t$ is the speed error to the future target speeds at $t' = t + \tau\Delta t$, where $\tau=5$ introduces an arbitrary delay. The final speed is used as target for the final $\tau{\Delta t}$ of the prediction horizon. The target speeds are retrieved from the predicted motion profile $\hat{\mathbf{s}}^i$, in particular we predict at 1 Hz and linearly interpolate them at 10 Hz. We cap acceleration using a maximum acceptable absolute acceleration $max_a=6.0\,m/s^2$ and absolute jerk $max_j=10.0\,m/s^3$ limits. Similarly, we use the orientation error for computing the steering angle input.
The orientation error ${\theta_\epsilon}^i_t$ is the difference between the current vehicle orientation and the target orientation, i.e. the orientation the vehicle would have if it was pointing directly to the goal position on the path. We then compute the curvature of the circle that the vehicle would describe as shown in Equation~\ref{eq:steering_angle_cap}.
\begin{equation}
\label{eq:steering_angle_cap}
\begin{aligned}
\kappa&= 2 \frac{\sin({\theta_\epsilon}^i_t)}{l_d}\\
\sigma^i_t&= \taninv(\kappa L)
\end{aligned}
\end{equation}

\subsection{Bayesian Inference}
The remaining characteristics of our system are \textit{i.)} processing the history of observations and \textit{ii.)} performing multimodal trajectory prediction. The first of these is important since individual observations can be noisy, while the latter is necessary to handle the uncertainty due to the unknown intention of the target agent. We achieve both by recursively consuming the observations in the history input and estimating the latent goal $g^i_t \in \mathbf{g}^i_t$ of the target agent $i$ via online Bayesian inference, see Equation~\ref{eq:bayes}. $P(g^i_{t=k-1} | \mathbf{y}^i_{t=k-1})$ represents the previous posterior and is used as a prior in the current time step. At $t=0$ a uniform prior $U(g^i_{t=k-1})$ is used.

\begin{equation}
\label{eq:bayes}
\hat{P}(g^i_{t=k} | \mathbf{y}^i_{t=k}) \propto P(\mathbf{y}^i_{t=k} | g^i_{t=k-1}) P(g^i_{t=k-1} | \mathbf{y}^i_{t=k-1})
\end{equation}

Our implementation relies on a single trajectory for a specific goal at a certain time step. The likelihood of each goal and its corresponding trajectory is estimated by comparing previously predicted trajectories $\hat{\mathbf{y}}^i_{t=k-1}$ with the observed agent state at current time step $t=k$. We extract the velocity direction $\phi^i_t$ from the velocity vector $\mathbf{v}^i_t$ components and use it in our likelihood estimation.
Variations between predicted states and observed states are captured assuming a normal distribution on the position and velocity direction of the agent at the current time step with fixed variances $\sigma^2$ for each term. We used the values $\sigma_x=0.4$, $\sigma_y=0.4$ and $\sigma_\phi=0.15$.

\begin{multline}
\label{eq:likelihood}
P(\mathbf{y}^i_{t=k} | g^i_{t=k-1}) \approx \mathcal{N}(x^i_{t=k} | \hat x^i_{t=k}, \sigma_x^2) \\ \cdot \mathcal{N}(y^i_{t=k} | \hat y^i_{t=k}, \sigma_y^2) \cdot \mathcal{N}(\phi^i_{t=k} | \hat \phi^i_{t=k}, \sigma_{\phi}^2)
\end{multline}

In certain situations, goals can only be reached if an agent executes an uncomfortable maneuver which should imply that those goals are less likely. To capture this aspect, trajectories with lateral absolute accelerations $\alpha^i_{t}$ above a predefined threshold $max_\alpha=0.0$ are penalised proportionally to the amount of violation. We integrate this as a bias term and multiply the probability of the corresponding goal with a weight that is computed using the maximum lateral acceleration $max_\alpha^{g^i}$ of the trajectory and an exponential decay function. See Equation~\ref{eq:lat_acc_bias}, where $\eta$ is a normalisation factor and $\lambda=0.5$ is the parameter that determines the amount of punishment. Lateral acceleration values are computed as $\alpha^i_{t} = \frac{{v^i_t}^2}{r^i_t}$ where ${r^i_t}$ is the radius of the circle that the vehicle is currently describing given its current steer angle and assuming a kinematic bicycle model.
\begin{equation}
\label{eq:lat_acc_bias}
\bar{P}(g^i_{t=k} | \mathbf{y}^i_{t=k}) = \eta \hat{P}(g^i_{t=k} | \mathbf{y}^i_{t=k}) \cdot e^{-\lambda (max_\alpha^{g^i} - max_\alpha)}
\end{equation}

An agent's goal may change in time. For example, one agent has finished a change-lane and wishes to perform a change-lane back in order to complete an overtake. Similarly to~\cite{baker2009action}, we add a forgetting step which has the effect of smoothing the posterior, balancing recent evidence and past evidence. We mix the output of the Bayes update with a uniform distribution to get the final posterior, see Equation~\ref{eq:forget}, where the parameter $\gamma=0.1$ determines the amount of smoothing.
\begin{equation}
\label{eq:forget}
P(g^i_{t=k} | \mathbf{y}^i_{t=k}) = (1 - \gamma) \bar{P}(g^i_{t=k} | \mathbf{y}^i_{t=k}) + \gamma U(g^i_{t=k})
\end{equation}

Finally, we need to account for goals changing due to the agent motion through the layout which will result in a change in the number of hypothesised goals. If a goal is no longer achievable, e.g. the agent has passed the exit ramp, then the goal is removed, and its mass is distributed equally among the remaining goals. Similarly, a new goal is added with mass equal to the value that it would have if we assume a uniform distribution over the complete set of hypothesised goals. This new mass $P(g_{new}) = \frac{1}{goals\,count}$ is equally drawn from the masses of the other already existing goals. Otherwise, there is a 1:1 mapping between goals at $t=k$ and those at $t=k-1$ and our map definition makes it straightforward to perform the matching.%

\section{EXPERIMENTS}
\label{experiments}
We evaluate our system on the Next Generation Simulation (NGSIM) dataset~\cite{colyar2007us101}, which is seen as a standard for highway scenarios. NGSIM includes vehicle trajectory data acquired from two US highways, US-101 and I-80, using CCTV cameras and a semi-automatic annotation process. Each dataset part was captured at 10\,Hz over
a time span of 45\,min and consists of 15\,min
segments of mild, moderate and congested traffic conditions. The dataset provides the coordinates of vehicles in UTM and a local reference system. We used UTM coordinates for alignment with our geo-referenced OpenDrive map annotations.
As in previous related works~\cite{deo2018convolutional}, we split the datasets into train (70\,\%), validation (10\,\%) and test (20\,\%) based on the vehicle ID. Each vehicle from each split is chosen as the target vehicle, defining one sample. We split the trajectories of the target vehicle into segments of 8\,s, where we use 3\,s of history and a 5\,s prediction horizon. The system hyper-parameters were tuned using randomised search on the training set.

\subsection{Overall System Evaluation}
We compare our system with other multimodal prediction methods using two standard trajectory error metrics, Root Mean Squared Error (RMSE) and Final Displacement Error (FDE). As in~\cite{mercat2020multi}, they are calculated by comparing the ground truth trajectory with the most likely (highest probability) trajectory. Lower scores are better. Table~\ref{methods-comparison} includes the numerical comparison. Both RMSE and FDE scores of our system are lower than that of a simpler baseline, the Constant Velocity (CV) model~\cite{mercat2019kinematic}, as well as that of other deep learning methods, CSP(M)~\cite{deo2018convolutional} and PiP~\cite{song2020pip}, for all time horizons. FDE for PiP is not reported in the original paper. We outperform the closest competitor SAMMP~\cite{mercat2020multi} for most time horizons (3\,s, 4\,s, 5\,s). Our system is comparable to the best state-of-the-art for shorter and easier horizons (1\,s, 2\,s, 3\,s) and significantly improve over the longer, more difficult horizons (4\,s, 5\,s), needed for motion planning~\cite{albrecht2021interpretable}.
\begin{table}[t]
\caption{Overall system comparison on the NGSIM test set using RMSE and FDE of the most likely trajectory.}
\label{methods-comparison}
\begin{center}
\begin{tabular}{c|c| c c c c c}
\hline
\multicolumn{2}{c|}{Time Horizon} & 1 s & 2s & 3s & 4s & 5s \\
\hline
\multirow{5}{*}{RMSE [m]} & CV~\cite{mercat2019kinematic} & 0.76 & 1.82 & 3.17 & 4.80 & 6.70\\
                          & CSP(M)~\cite{deo2018convolutional} & 0.59 & 1.27 & 2.13 & 3.22 & 4.64\\
                          & PiP~\cite{song2020pip} & 0.55 & 1.18 & 1.94 & 2.88 & 4.04\\
                          & SAMMP~\cite{mercat2020multi} & \textbf{0.51} & \textbf{1.13} & 1.88 & 2.81 & 3.98\\
                          & Flash & \textbf{0.51} & 1.15 & \textbf{1.84} & \textbf{2.64} & \textbf{3.62}\\
\hline
\multirow{4}{*}{FDE [m]} & CV~\cite{mercat2019kinematic} & 0.46 & 1.24 & 2.27 & 3.53 & 4.99\\
                          & CSP(M)~\cite{deo2018convolutional} & 0.39 & 0.91 & 1.55 & 2.36 & 3.39\\
                          & SAMMP~\cite{mercat2020multi} & \textbf{0.31} & \textbf{0.78} & 1.35 & 2.04 & 2.90\\
                          & Flash & 0.33 & 0.82 & \textbf{1.34} & \textbf{1.91} & \textbf{2.61}\\
\hline
\end{tabular}
\end{center}
\end{table}

\subsection{Motion Profile Prediction Analysis}
The previously reported overall error can be caused by several factors. In this section, we focus on what we observed to be the most significant contributor: motion profile prediction. In addition to the previously reported dataset pre-processing steps, we split the dataset based on the observed behaviour of the agent being predicted: follow-lane and change-lane. We evaluate the motion profile prediction component using the RMSE and Mean Negative Log Likelihood (MNLL) errors on the predicted future distance. MNLL takes uncertainty into account~\cite{deo2018convolutional}. We adapt the RMSE to compute it on a single dimension. The NLL is already defined in Equation~\ref{nll_loss}, where the displacement error at time $t$ for a Gaussian component $m$ centered at $\boldsymbol{\tilde{d}}_{m, t}$ is $\boldsymbol{\epsilon}_{m, t} = \mathbf{d}_{m, t} - \tilde{\mathbf{d}}_{m, t}$. We average across the dataset to compute the MNLL value.

\begin{figure}[t]
  \centering
  \includegraphics[width=0.8\columnwidth]{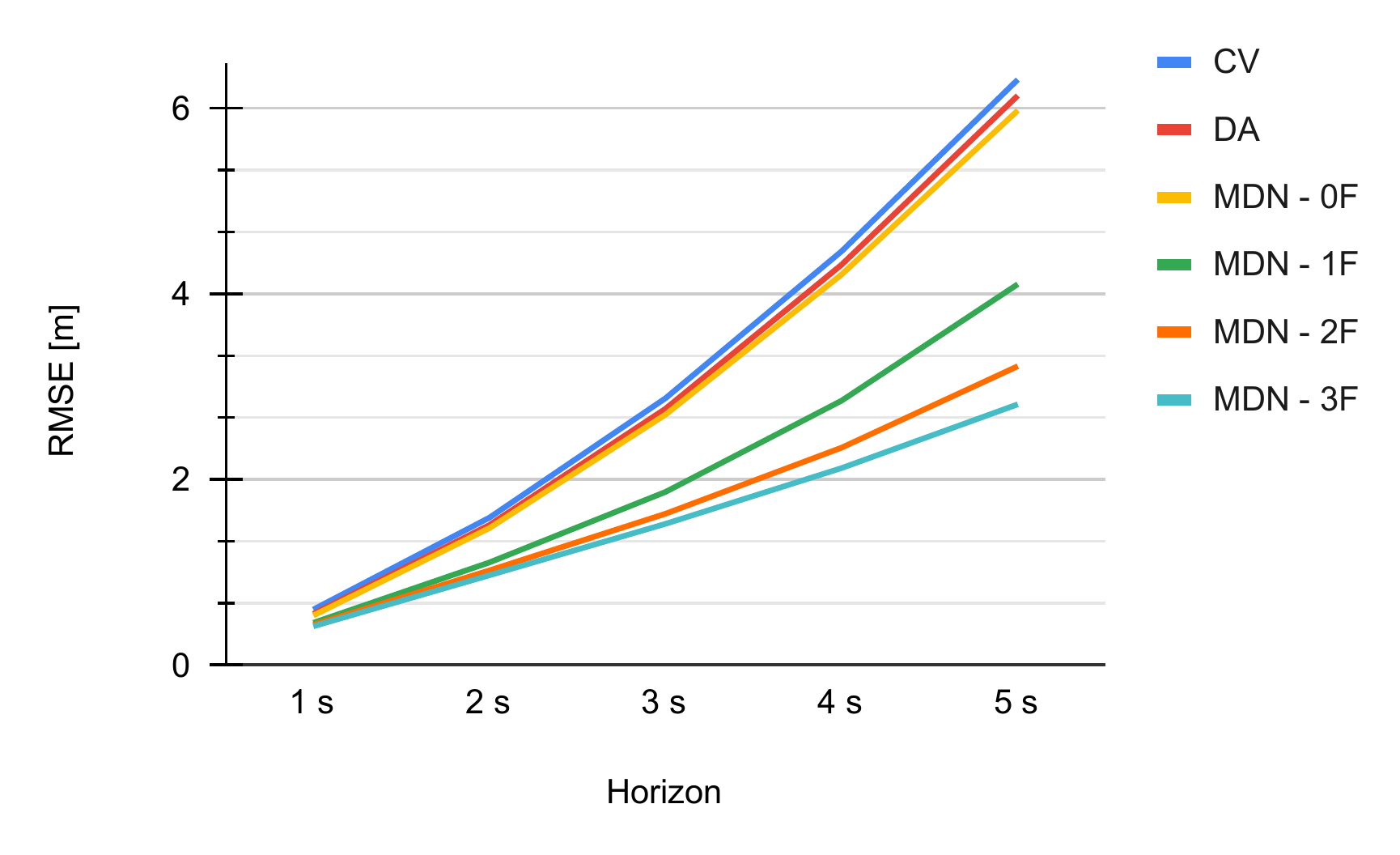}
  \includegraphics[width=0.8\columnwidth]{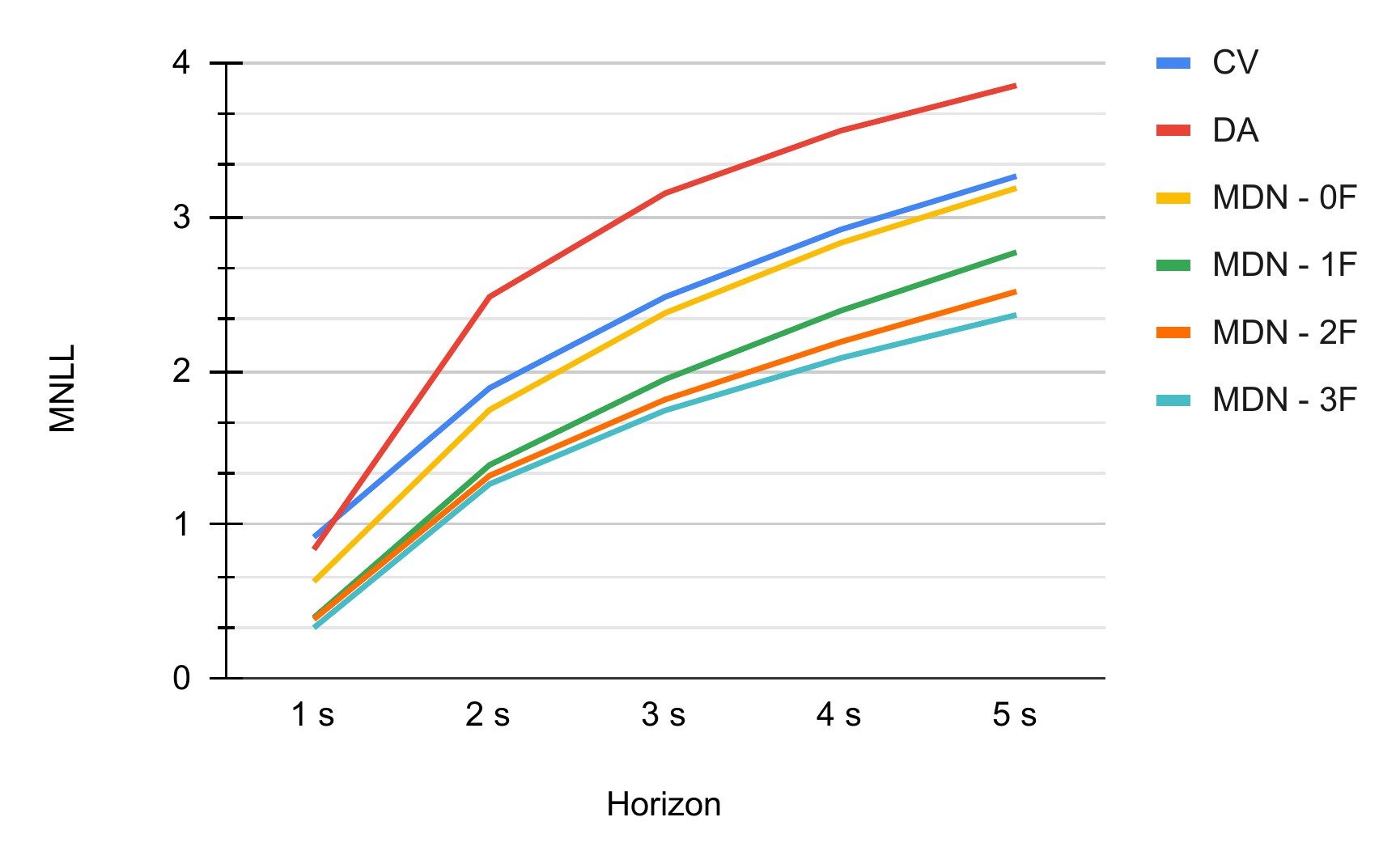}
  \caption{RMSE and MNLL comparison of follow-lane models for motion profile prediction. Models considering more front agents are more accurate.}
  \label{fig:lane_follow_mp_comparison}
\end{figure}

Figure~\ref{fig:lane_follow_mp_comparison} shows the relative performance of different follow-lane motion profile prediction models. We compared each neural network in the collection with physics-based methods: Constant Velocity (CV) and Decaying Acceleration (DA).
We model the physics-based model uncertainty at each predicted time step using standard deviations and modelling the errors of the CV or DA assumption with a centered Gaussian distribution at each time step. Neural networks outperform physics-based models. Considering more agents leads to lower errors since traffic dynamics, such as safe distancing, stop and go motion and motion initiation, can be modelled. For example, we observed that considering 3 front agents instead of 0 reduces motion initiation errors; RMSE and MNLL at 5\,s decrease by 3.21\,m and 0.50 respectively.

We analysed the relative performance of different change-lane motion profile prediction models. We report the weighted RMSE and MNLL at 5\,s for brevity. We averaged the performance on each dataset split, varying the number of front and side agents. The collection (RMSE 7.14\,m, MNLL 3.44) always improves over physics-based models CV (RMSE 10.09\,m, MNLL 4.02) and DA (RMSE 9.81\,m, MNLL 3.84). In addition, Tables~\ref{tab:lane_change_mp_comparison_rmse} and \ref{tab:lane_change_mp_comparison_nll} report the performance of each change-lane network in the collection for further analysis. Even though the attention of a driver performing a change-lane should be on what happens in both the current and target lane, we observe that the performance of the change-lane networks is mainly influenced by the number of side agents. The performance of the change-lane networks is not influenced by the number of front agents as much as the performance of the follow-lane networks.
\begin{table}[t]
\caption{Comparison of change-lane motion profile prediction models on the NGSIM test set using RMSE at 5\,s}.
\label{tab:lane_change_mp_comparison_rmse}
\begin{center}
\begin{tabular}{|l|*{4}{c|}}\hline
\backslashbox{\# Front}{\# Side}
& \multicolumn{1}{c|}{0} & \multicolumn{1}{c|}{1} & \multicolumn{1}{c|}{2} & \multicolumn{1}{c|}{3}\\\hline
0 & 10.05 & 7.54 & 6.45 & 5.13\\\hline
1 & 9.49 & 7.71 & 6.44 & 5.13\\\hline
2 & 9.50 & 7.05 & 6.15 & 5.06\\\hline
3 & 8.98 & 8.09 & 6.14 & 5.25\\\hline
\end{tabular}
\end{center}
\end{table}
\begin{table}[t!]
\caption{Comparison of change-lane motion profile prediction models on the NGSIM test set using MNLL at 5\,s.}
\label{tab:lane_change_mp_comparison_nll}
\begin{center}
\begin{tabular}{|l|*{4}{c|}}\hline
\backslashbox{\# Front}{\# Side}
& \multicolumn{1}{c|}{0} & \multicolumn{1}{c|}{1} & \multicolumn{1}{c|}{2} & \multicolumn{1}{c|}{3}\\\hline
0 & 3.82 & 3.44 & 3.24 & 3.04\\\hline
1 & 3.83 & 3.52 & 3.39 & 3.00\\\hline
2 & 3.74 & 3.34 & 3.33 & 3.01\\\hline
3 & 3.56 & 3.77 & 3.68 & 3.28\\\hline
\end{tabular}
\end{center}
\end{table}
The most difficult cases for all change-lane models, including physics-based models, involve fewer number of side agents. Our interpretation is that the number of ways one can perform a change-lane is reduced in congested situations, simplifying the prediction task. Furthermore, the RMSE values vary a lot with number of side agents in comparison to the MNLL values. The cases with few agents have more variability, which is captured by our system handling of uncertainty as confirmed by the MNLL values.

\subsection{Qualitative Analysis}
One major advantage of our system is the ability to inspect each component, allowing to debug and understand their contributions to the overall performance.
Here, we show how we can debug and interpret the output of the Bayesian inference component. We also provide insight in the advantages of combining the neural networks with a classical trajectory generator.

\begin{figure}[t]
  \centering
  \subfloat[\centering Time 0.0]{{\includegraphics[scale=0.25, trim={6cm 6cm 1cm 2.5cm}, clip]{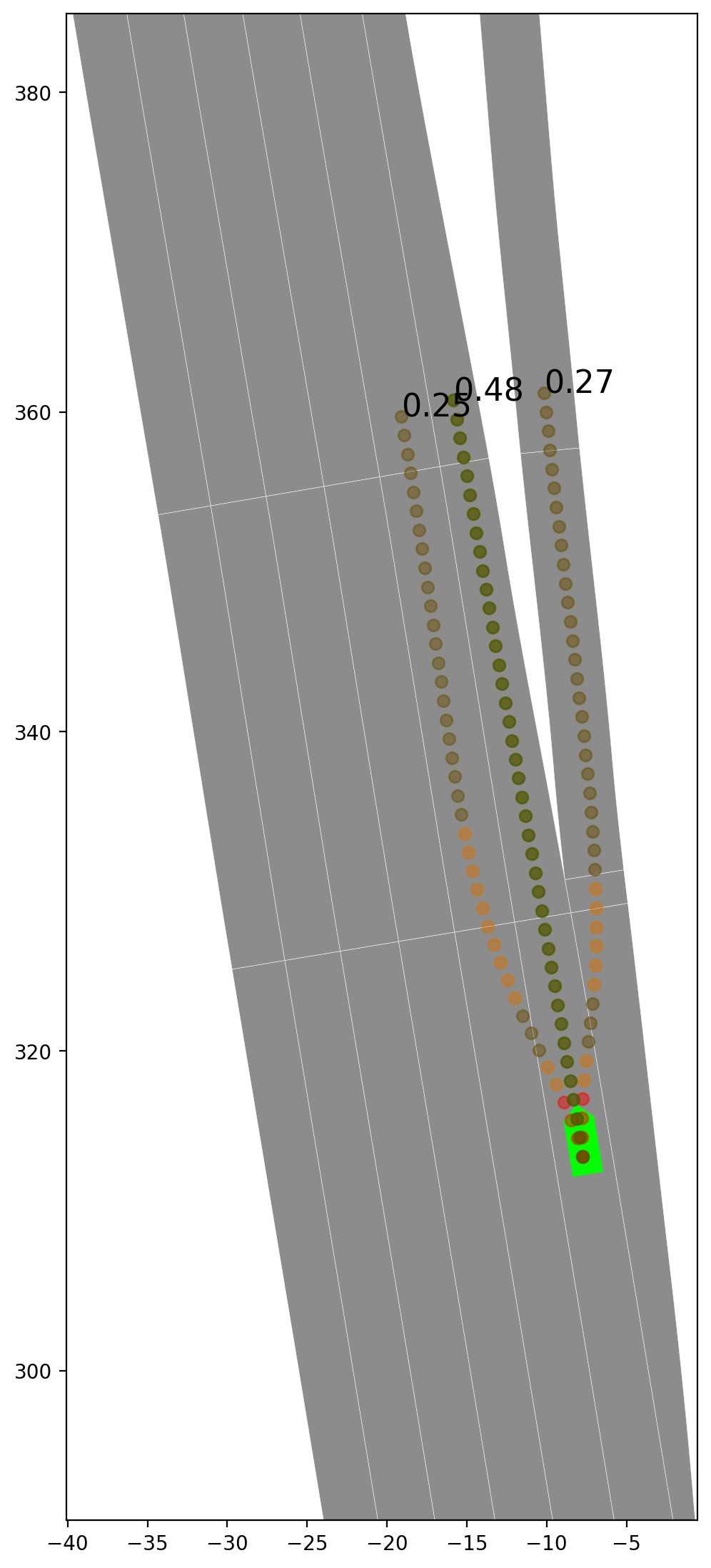} }}%
  \qquad
  \subfloat[\centering Time 0.2]{{\includegraphics[scale=0.25, trim={6cm 6cm 1cm 2.5cm},clip]{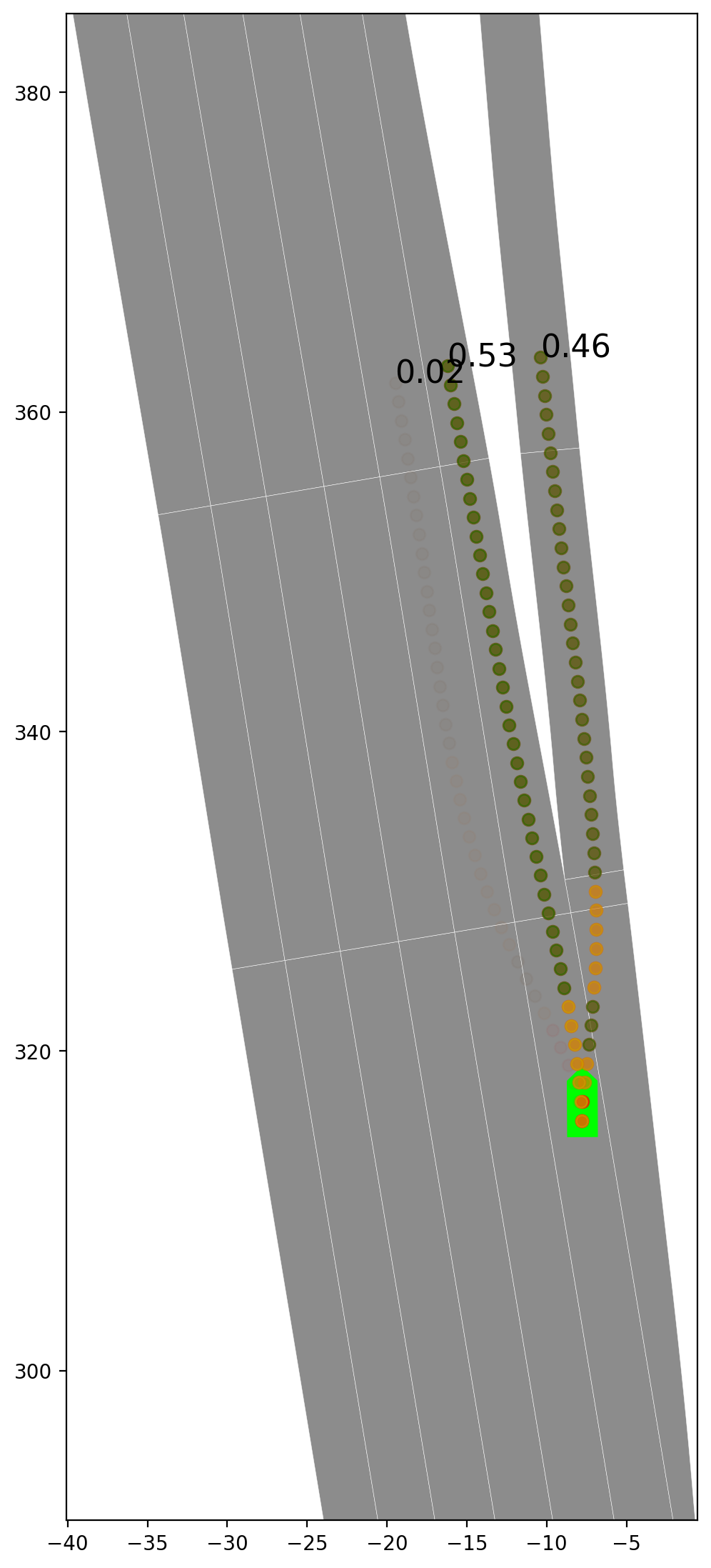} }}%
  \qquad
  \subfloat[\centering Time 0.4]{{\includegraphics[scale=0.25, trim={6cm 6cm 1cm 2.5cm}, clip]{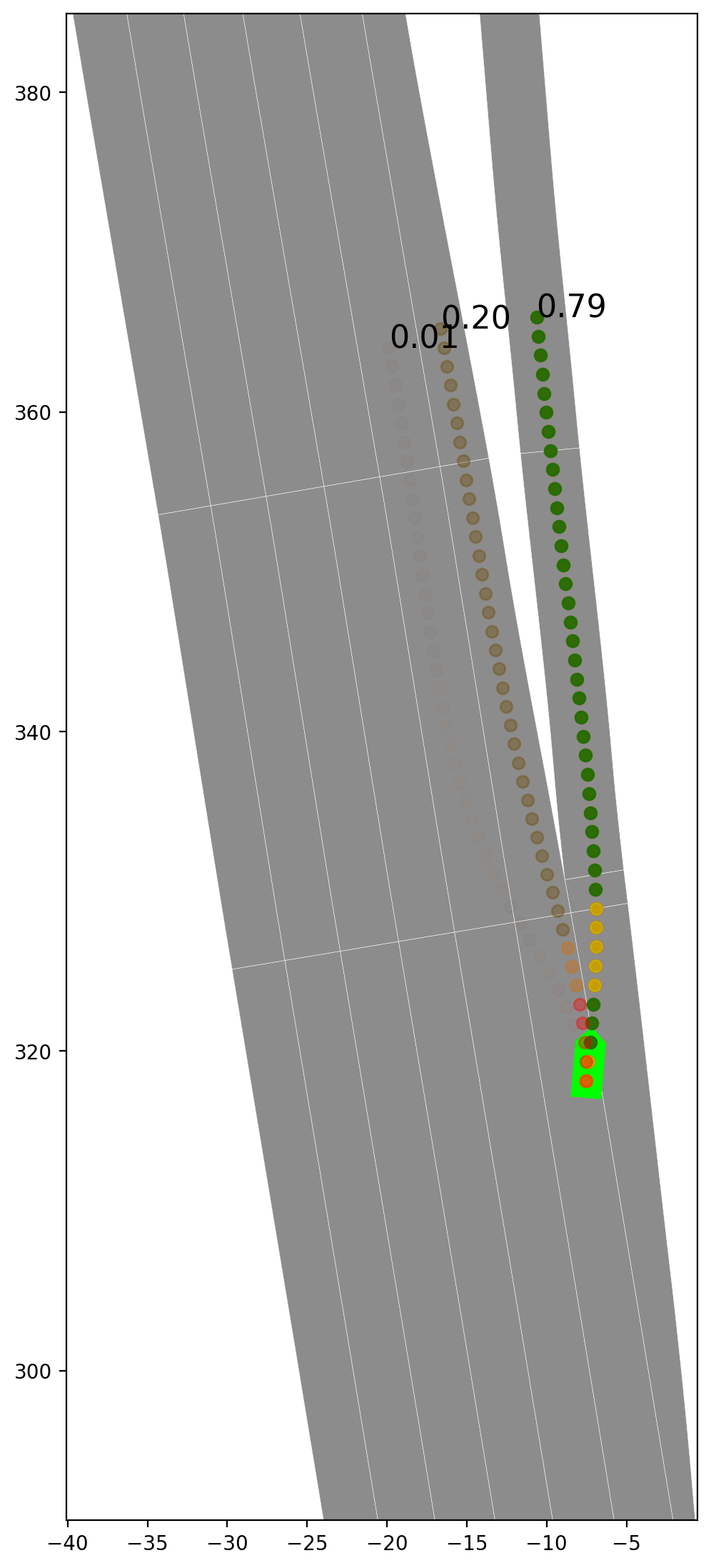} }}%
  \qquad
  \subfloat[\centering Time 0.6]{{\includegraphics[scale=0.25, trim={6cm 6cm 1cm 2.5cm}, clip]{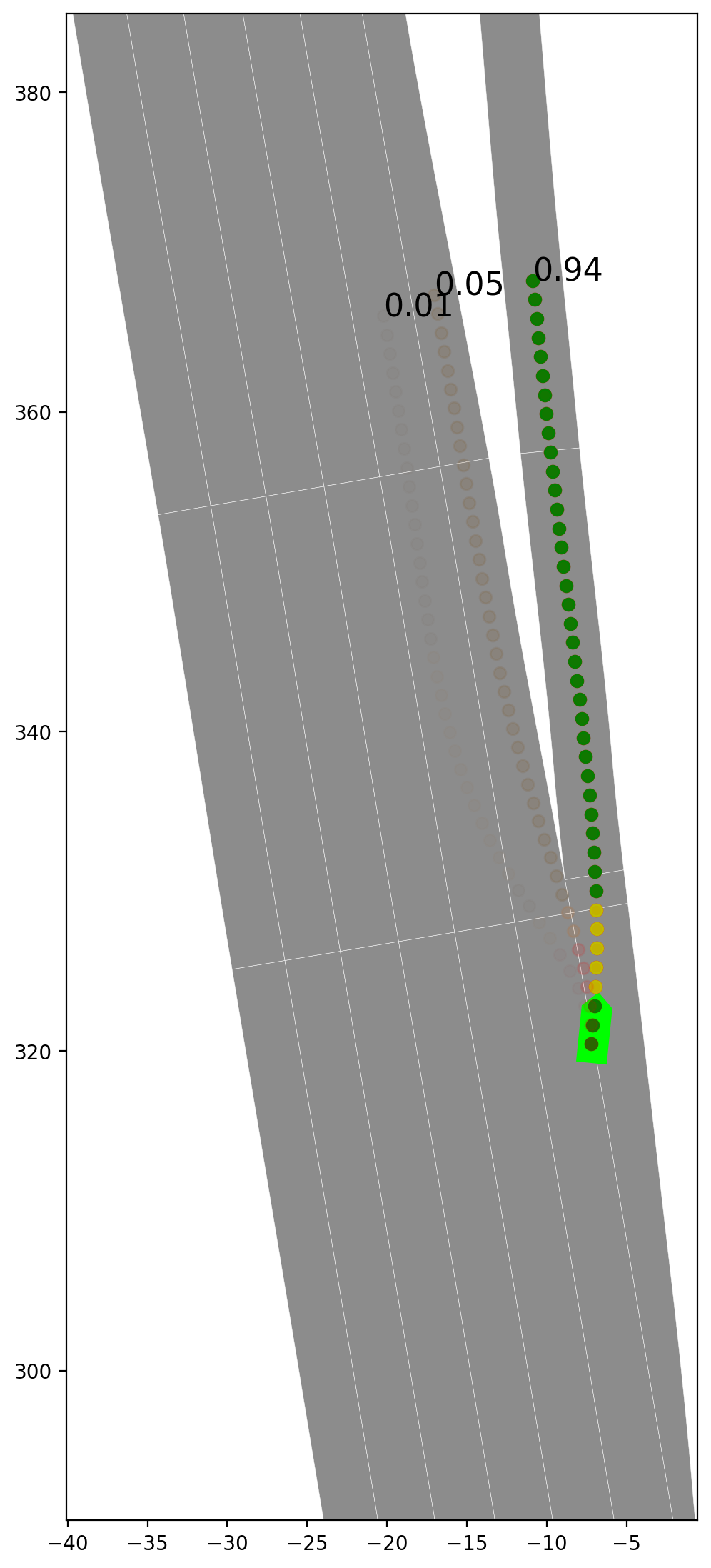} }}%
  \caption{Bayesian inference component on the I-80 highway exit with a ${\Delta t}=0.1$ but showing images with a ${\Delta t}=0.2$. The green box represents the observed position and orientation of the vehicle. The sequences of points starting from the vehicle are the predicted trajectories, and these are annotated with their posterior probability. The colour of the points represent the lateral acceleration: green is low ($<2.0$), yellow is medium ($<5.0$) and red is high ($\geq5.0$). Transparency is proportional to the probability of trajectories.}
  \label{fig:bayesian_inference_analysis}
\end{figure}

We illustrate the Bayesian inference process of inferring the goal of the target agent in Figure~\ref{fig:bayesian_inference_analysis}. For ease of visualisation, we use a constant velocity motion profile prediction model and we do not show the neighbour agents as this context information is handled in the trajectory generator component. In the first observation, the most likely behaviour is a follow-lane as the change-lane-left and exit-right contain high lateral acceleration values, reducing their likelihood. After a couple of observations, the vehicle has turned towards the exit, causing the change left to be predicted as highly unlikely. Follow-lane is still very probable as the vehicle has not gone far from the centre of the lane. After another pair of observations, the follow-lane becomes unlikely due to the large lateral acceleration values. The final observation results in a large probability in the exit behaviour, while the other behaviours have low values. The forgetting factor aids in quickly responding in the event of a goal change as these values do not go below a minimum.

In addition to the known unpredictability of neural network performance~\cite{angelos2020}, the training data can be noisy. Due to this combination, their output is likely to violate our acceleration and jerk limits. To illustrate the impact of these limits, we measure the number and magnitude of violations. The predicted trajectory closest to the ground truth trajectory is used for these measurements. On the motion profile validation dataset made of approximately 840k samples, we observed 38 acceleration violations and 241k jerk violations that were corrected by the limits. The mean of the acceleration violation is $2.52\,m/s^2$ with standard deviation $1.86\,m/s^2$. The mean of the jerk violation is $12.30\,m/s^3$ with standard deviation $8.01\,m/s^3$. We also noticed that introducing the limits did not significantly impact the overall performance of the system.

\begin{figure}[t]
  \centering
  \includegraphics[scale=0.22]{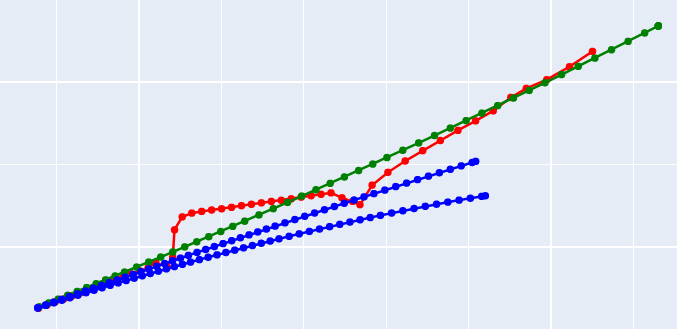}
  \caption{Violations of kinematic and motion profile constraints. The ground truth trajectory in red presents high curvature and large variations in point distances. The predicted trajectories in blue and green (most likely) are well-formed.}
  \label{fig:violation}
\end{figure}

Previous work~\cite{girase2021physically} has also reported that the path output of neural networks can violate the constraints of a vehicle model. These concerns are important since a prediction system that produces infeasible trajectories can negatively impact the performance of the components down the line, e.g. planning. Figure~\ref{fig:violation} shows that the combination of neural networks with kinematic and motion profile constraints can address these concerns. Even though the ground truth trajectory is not feasible, the prediction output is. Furthermore, the green trajectory is the one with the highest probability.

\subsection{Runtime Analysis}
\begin{table}[t]
\caption{Running times [ms] per component call.}
\label{tab:runtime_analysis}
\begin{center}
\begin{tabular}{{|c|c|c|}}\hline
Data processing & Motion Profile Prediction & Bayesian Inference \\\hline
3.3 & 1.8 & 0.4\\\hline
\end{tabular}
\end{center}
\end{table}
The inference runtime of the system was measured with a desktop PC equipped with an Intel Core i7-7800X CPU 3.50GHz over a set of 1000 samples chosen arbitrarily from the dataset. Our full system which runs on CPU is implemented in C++ and Python. It takes approximately 5.5\,ms per call for each agent and we provide a breakdown of the components cost in Table~\ref{tab:runtime_analysis}. Given our code structure, the time reported for the Bayesian inference component includes the time for generating trajectories with the pure pursuit algorithm. The feature extraction step is currently the bottleneck as it is Python code which can be further optimised. Training the mixture of experts collection in Tensorflow requires approximately 32.5 minutes on the full training dataset on CPU while other methods report training times of several hours using the same dataset and high-end GPUs, e.g. 17.5 hours~\cite{mersch2021maneuver} or 1 day~\cite{tang2019multiple}.

\section{CONCLUSIONS}
\label{conclusions}
We presented a novel motion prediction system, based on a modular architecture involving both data-driven and analytically modelled components. We showed that it achieves state-of-the-art results in the highway driving setting. The proposed system covers multiple aspects of interest, namely multi-modality, physical feasibility, motion profile uncertainty, system maintainability and efficiency. %

\bibliography{mybib}{}
\bibliographystyle{ieeetr}

\end{document}